\documentclass[
  english,        
  font=times,     
  onecolumn,      
]{article}

\usepackage{lineno}
\usepackage{placeins}
\usepackage{amsmath}
\usepackage{csquotes}
\usepackage{float}
\usepackage{booktabs}
\usepackage{multirow}
\usepackage{multicol}
\usepackage{siunitx}
\usepackage[utf8]{inputenc} 
\usepackage{colortbl} 
\usepackage{chemformula} 
\usepackage{xr} 
\makeatletter
\newcommand*{\addFileDependency}[1]{%
  \typeout{(#1)}%
  \@addtofilelist{#1}%
  \IfFileExists{#1}{}{\typeout{No file #1.}}%
}
\makeatother

\newcommand*{\myexternaldocument}[2][]{%
  \externaldocument[#1]{#2}%
  \addFileDependency{#2.tex}%
  \addFileDependency{#2.aux}%
}
\myexternaldocument{SI}

\DeclareSIUnit\angstrom{\text {Å}}
\DeclareSIUnit\rydberg{\text{Ry}}
\DeclareSIUnit\hartree{\text{Ha}}
\DeclareSIUnit\bohr{\text {\ensuremath {a}}_{0}} 

\usepackage{geometry}
\usepackage{array} 
\usepackage{caption} 
\usepackage{makecell} 
\usepackage{hyperref}
\usepackage{float}      
\usepackage{chemformula} 
\usepackage{tabularx}   
\usepackage{placeins}   
\usepackage{graphicx}   
\usepackage{caption}    
\usepackage{amsmath}    
\usepackage[capitalise, nameinlink]{cleveref} 

\usepackage{comment}
\usepackage{bm}
\usepackage[natbib=true,
    backend=bibtex,
    style=numeric,
    sorting=none
]{biblatex}

\usepackage{etoolbox}

\title{PYTHIA, a transferable machine learning framework for simultaneous defect and adsorption site optimization}

\author{
  Ioannis Kouroudis$^{1,2}$\thanks{ioannis.kouroudis@tum.de}
  \and
  Zhaosu Gu$^{1,2}$
  \and
  Poonam$^{2}$
  \and
  Leon Müller$^{3}$
  \and
  Prof.\ Dr.\ rer.\ nat.\ Alessio Gagliardi$^{2}$%
  \thanks{alessio.gagliardi@tum.de}
}

\date{ }

\begin{document}
\maketitle
\begin{center}
\small
$^{1}$ These authors contributed equally

\vspace{0.5em}

$^{2}$ Chair of Simulation of Nanosystems for Energy Conversion,
Department of Electrical Engineering, TUM School of Computation,
Information and Technology, Atomistic Modeling Center (AMC),
Munich Data Science Institute (MDSI),
Technical University of Munich,
Hans-Piloty-Straße 1, 85748 Garching, Germany

\vspace{0.5em}

$^{3}$ Fritz-Haber-Institut der Max-Planck-Gesellschaft, Faradayweg 4-6, Berlin, Germany
\end{center}

\begin{abstract}
Machine learning has permeated material optimization, providing high value. Nevertheless, its reliance on a high number of samples limits its applicability to relatively explored material families. In this paper we extend the applicability of machine learning to novel and understudied materials through the PYTHIA (Probabilistic Yet Transferable Hierarchical Interaction Architecture) framework. Pythia utilizes two of the most powerful machine learning concepts, foundational models and transfer learning. We employ a complex foundational model trained on a large dataset of chemically similar molecules, on a target that is physically close to our prediction target. Then, we freeze the weights and perform subspace alignment with the use of a transformer layer and add uncertainty quantification with gaussian processes. These two layers can be trained on a limited dataset yet allow us to leverage the gained knowledge and the studied materials and complex architectures. The resulting model is used as a surrogate for a Bayesian Optimization scheme. To showcase the robustness of the framework, we applied it to the investigation of adsorption and defect optimization of two distinct systems. Not only were we able to correctly and efficiently identify the optimal adsorption sites of even complex adsorbates, the algorithm showed remarkable accuracy even for systems not present in its training set. 
\end{abstract}

\section{Introduction}
Adsorption plays an important role in the interaction of nanostructured materials with molecules and ions. These interactions depend on the types of atoms present and their local arrangement. 
Identifying stable adsorption structures also helps explain electrochemical reactions at metal surfaces \cite{Sklason2007}. For materials with different compositions or defects, this search becomes more difficult. Changes in the local atomic environment can affect both the preferred adsorption position and the strength of binding. To investigate this, Density functional theory (DFT) is widely used. Mapping the potential energy surface (PES) requires calculations for many molecular positions and orientations \cite{Wei1998a}, a search that can be automated by various tools which help generate, screen, and relax these structures \cite{Marti2021, Pedretti2023}. Other methods examine different parts of the energy landscape. Minima hopping and evolutionary algorithms search for structures with low energies \cite{Goedecker2004, Schnborn2009}. The nudged elastic band (NEB) method finds minimum energy paths between two energy states \cite{jonsson1998nudged}, while metadynamics explores free energy landscapes \cite{barducci2008well, YazdanYar2018}. These methods provide useful information, but their computational cost can remain high when each energy evaluation requires a DFT calculation. The cost increases further when several substrate compositions and defect structures are studied.

Machine learning can reduce the number of expensive calculations. It has been applied to many problems in materials research \cite{mayr2022machine,lampe2023rapid,kouroudis2023utilizing}. In atomistic simulations, machine learning models can describe surface reaction PESs \cite{Lorenz2006} and predict adsorption energies on nanoclusters \cite{jager2018machine}. Graph neural networks have also been used to predict molecular adsorption energies on metals \cite{PabloGarca2023}, and learned interatomic potentials can speed up adsorption searches \cite{Lan2023}. However, these models need suitable training data. For a new adsorption system, generating enough DFT data can account for much of the total computational cost, often times providing a full system investigation before the model is even trained.

Active learning (AL) and Bayesian optimization (BO) help address this problem by using the available data to choose new calculations. Bayesian models can learn atomic interactions during simulations \cite{Jinnouchi2019, Xie2023}, and uncertainty estimation is also an active area of research for molecular models \cite{wollschlager2023uncertainty}. Related approaches have been used for crystal structure prediction \cite{Yamashita2018}, the optimization of nanoporous materials \cite{Deshwal2021a}, materials discovery with graph models \cite{Zuo2021}, and catalyst screening through adsorption calculations \cite{Tran2018a}. Machine learning can also accelerate searches for saddle points and reaction paths \cite{Peterson2016, Koistinen2019}, as well as stable adsorbate geometries \cite{Jung2023}. 

The choice of structural representation affects how well these methods work. BO has been used to search for molecular conformers by varying torsion angles \cite{Chan2019}. The Bayesian Optimization Structure Search (BOSS) framework uses structural coordinates to explore energy landscapes \cite{todorovic2019bayesian}. Its applications include molecular adsorption on surfaces \cite{Jaervi2020} and the search for molecular conformations on metal clusters \cite{Fang2023}. These coordinates are useful when the main degrees of freedom can be identified in advance. However, a representation designed for one system may need to be changed when the molecular composition, substrate, or defects change. Learning a representation directly from atomic structures offers a way to handle these differences.

In our previous work, we developed AUGUR to identify adsorption sites by combining a graph neural network, Gaussian process (GP) regression, and BO \cite{kouroudis2025augur}. AUGUR learns a structural representation through a graph neural network (GNN) and uses it to predict energies and estimate uncertainty via a GP, without requiring manually designed molecular descriptors. 
A very relevant question however is, how this approach can benefit from knowledge learned on other atomic systems. Such knowledge could reduce the amount of data needed for a new adsorption problem. This is particularly useful when the search includes several compositions or defect types, each with only a limited number of reference calculations.

Pretrained atomic models provide a starting point for this transfer. The MACE architecture \cite{MACE} has been used to develop foundation models trained on many different structures \cite{batatia2025foundation}. The atomic features learned by these models can then be used in new applications. However, pretraining does not guarantee accurate adsorption energies for every system, especially when its chemical environments or reference calculations differ from those used in training. Previous work has shown that adapting a pretrained potential while keeping some parameters fixed can reduce the amount of training data required \cite{radova2025frozen} and even extrapolate to new properties \cite{Kouroudis2026}. For adsorption searches, we also need a model that adapts to the available data and estimates uncertainty to guide the choice of new calculations.

To address the above concerns we introduce PYTHIA, which combines a pretrained MACE model, a Transformer, GP regression, and BO. The pretrained MACE  provides atomic features that contain relevant descriptors for an adjacent system and property. A Transformer \cite{vaswani2017attention} processes these features and produces a representation that has translated the features to the current system. This vector is then used as input for the GP that predicts the adsorption energy and its corresponding uncertainty. During training, the MACE parameters remain fixed, while the Transformer and GP are trained on data from the systems being studied. This allows the model to reuse features learned during pretraining and adapt them to the target energy prediction problem. BO then uses these predictions to select candidate defects and adsorption positions for further electronic structure calculations. This pipeline provides all the benefits of large and complex models, without the usual necessity of large datasets.

We verify PYTHIA on two highly relevant cases, optimizing both the defects of the cluster and the position of the adsorbate simultaneously. 
The first is the Zn adsorption on a defective Chini cluster. This system is of high interest due to its role in the chemistry of metal carbonyl clusters \cite{Mechanistic_Poonam} and their encapsulation properties. 
The second examines adsorption of formaldehyde on a buckled silicon layer \cite{Silicene_str, Silicene_corrugation}. Formaldehyde is a harmful chemical and silicene is a periodic and complex structure. Therefore the optimization of this system provides both a high environmental impact as well as a heightened complexity.
These systems provide different local chemical environments for deep testing of the model. We compare PYTHIA with two baseline models, AUGUR and a fully trained MACE with a GP as a final layer and was found to outperform both by a significant margin showing that it has higher complexity than AUGUR and lower data demand then the full MACE.  We also examine the predictions for defect groups not present in the training data and found our model highly accurate. 
To summarise, we show that PYTHIA combines the high accuracy of complex models at a fraction of the data cost. Additional to the predictions, PYTHIA also provides highly tailored optimization, identifying sites with many times better adsorption energy than present in the training set. With novel materials emerging fast, this data-efficient transferable framework can significantly expedite chemical investigation alleviating the need for time consuming data acquisition.


\section{Methods}
\subsection{PYTHIA pipeline}

\begin{figure}[h!]
    \centering
    \includegraphics[width=1\linewidth]{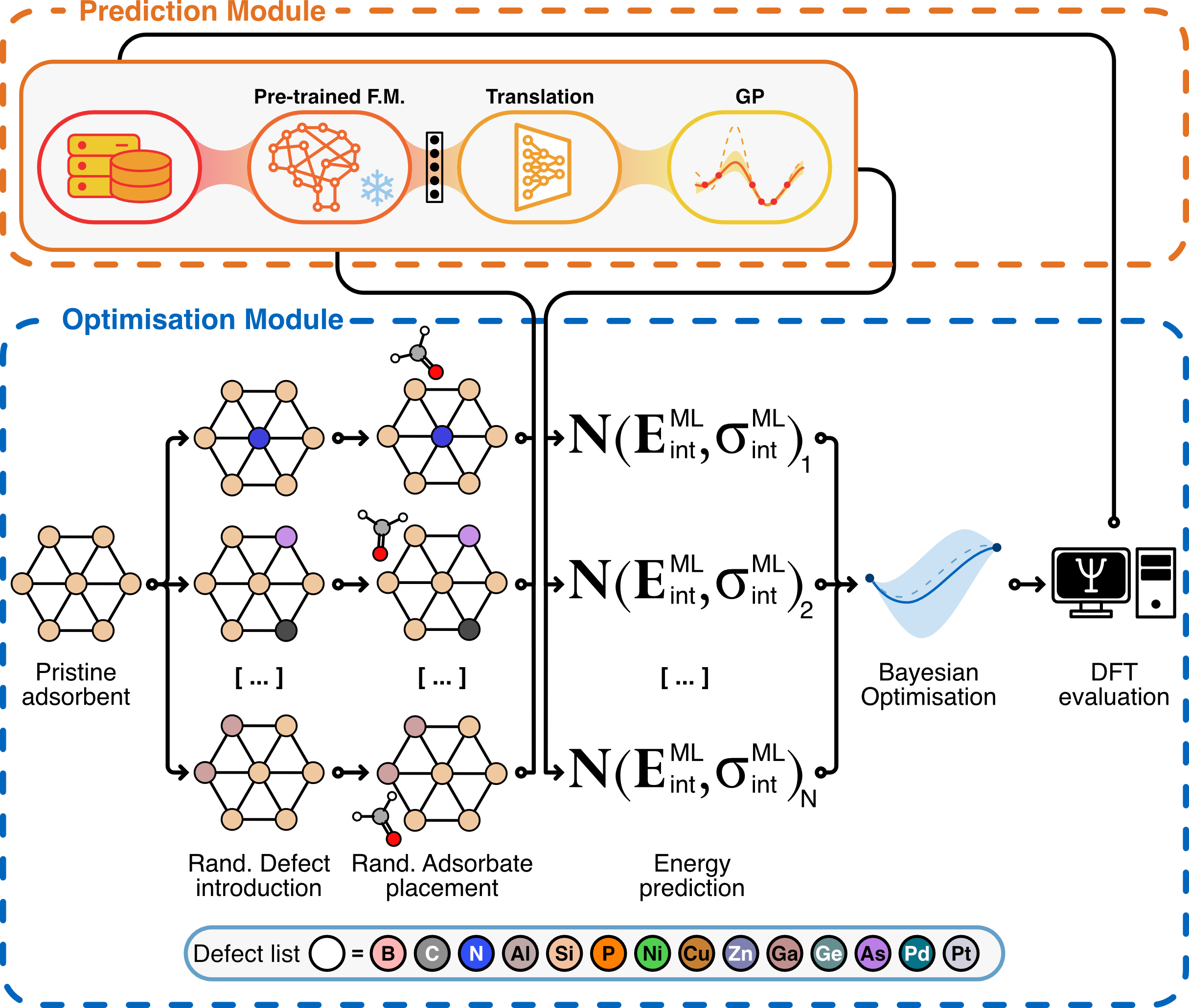}
    \caption{Pythia Pipeline. The top row corresponds to the predictive part of the framework while the bottom is the permutation of defects and placements, along with the optimization. Both together consitute the total optimization process. }
    \label{fig:pipeline}
\end{figure}
In \cref{fig:pipeline} the entire PYTHIA pipeline is shown for adsorption site optimization. 
It must be noted that the pipeline provides a prediction for the overall interaction energy surface and its corresponding uncertainty. The combination of those will allow the user to decide on an individual basis
whether the added computational cost of another DFT round offsets the potential uncertainty decrease of the predictions.

\FloatBarrier
\subsection{Bayesian Optimization}
Bayesian Optimization (BO) is an optimization algorithm that has proven itself to be both data efficient and accurate even in non-convex/concave optimization problems. 
It is especially useful in optimization problems where there is a costly underlying evaluation function, such as time consuming experiments or simulations. Data-driven models substitute these procedures with a comparatively inexpensive surrogate, that, however, requires data to be trained. Oftentimes, this data demand is high enough as to invalidate the eventual benefits. Therefore an efficient and data-sparse dataset is necessary to leverage the full advantage of machine learning. 
Bayesian optimization gives answers to this bottleneck, simultaneously optimizing both the underlying problem as well as the sampling one.
This is done by combining two components, a stochastic predictor, typically a GP regressor and a function that weighs each hopeful optimization point based on its optimality and amount of new information it brings into the model. This function is known as the acquisition function and can take many forms depending on the individual case needs. For the remainder of the chapter, we will describe a minimization process, but everything can be applied without loss of generality to any optimization problem. 
In the present work, we seek to find the position of adsorption and set defects that will result in the lowest possible energy. To this end, we can generate a number of random positions and evaluate their respective energies using the uncertainty-aware predictor.  The prediction's uncertainty is reflected in its standard deviation. Initially, we choose to evaluate points predicted to have high standard deviation, regardless of their predicted energy. In this way, we are giving the model information about compositions and sites far removed from the ones already in the training set. Once this process has sufficiently trained the model, the focus can shift on determining the systems that exhibit good adsorption energies, thus fulfilling our original optimization goal. These stages are called exploration and exploitation respectively.
Note that there are many ways to optimize the acquisition functions. The high number of physical constraints however made most of them not trivially applicable. To this end, we have chosen to generate 10.000 individual systems and predict them. Though this is slower than traditional gradient based approaches, it allows for the strict control of the evaluated system's physical feasibility. Further, PYTHIA is efficient enough for the evaluation to still be performed in negligible time frames.  More information about how the points were generated can be found in \Cref{chap:point_generation}.

\subsection{Point generation}\label{chap:point_generation}
The convergence of Bayesian Optimization can be significantly sped up if we introduce a layer of physical understanding of the problem. The simplest way is to generate a number of points (in our case, 10.000) that adhere to a set of constraints. 
A random number of defects and a corresponding number of random atoms of the cluster is chosen. These atoms are permuted, each by a unique, randomly selected element from the valid pre-defined defect elements. This has created our defective adsorbant. Subsequently, one random atom of the adsorbant and one random atom of the adsorbate is chosen. The chosen adsorbate atom is placed on the surface of a sphere with a radius determined by the physical limits of the interaction distance between the two atom types and centered on the chosen adsorbant atom. A random rotation matrix is defined and the adsorbate is rotated thus. The resulting adsorbate is tested to ensure that none of its atoms pass through or lie in the adsorbant. This is done by performing a Delaunay tessellation on the cluster coordinates and determining the location of the adsorbent with respect to the resultant convex hull coordinate comparison. If this constraint is violated, the system is discarded and the process restarts. If the system passes this check, the distance between the adsorbate and remaining adsorbant atoms is calculated. If through the rotations any atom is placed closer than physically possibly the system is discarded. Otherwise, the new system is added to the sample set and evaluated through the pipeline.  The sample generation process can be summarized as follows:

\begin{enumerate}
    \item Choose a random number of cluster atoms and exchange them for a set of random elements that are capable of producing valid, defective structures.
    \item Choose a random atom of the cluster, provided adsorption is physically possible on it.
    \item Place a random adsorbent on the surface of a sphere with a radius determined by physical knowledge of the maximum and minimum interaction distance of these two atoms. The functionality to determine a specific adsorbent atom for placing on the sphere is also provided.
    \item Rotate the adsorbent around its center of mass by a random set of angles.
    \item Test if the adsorbent either lies in or passes through the cluster. 
    \item If the above constraint is not violated, test if the adsorbent is closer than a distance of tolerance to any cluster atoms.
    \item If the above constraint is not violated, the sample is admitted into the list of viable samples.
\end{enumerate}

\FloatBarrier
\subsection{Gaussian Processes}
In the current project, the uncertainty quantifying model chosen was Gaussian processes as it combines robustness to overfitting and the well-documented accuracy of global kernel methods. It is a data-driven stochastic algorithm that models predictions as the posterior of the Bayes formula. The prior and the likelihood are modeled as Gaussian distributions whose parameters are optimized based on the already measured samples. In this way, the algorithm elegantly provides an inherent uncertainty quantification, which in data sets of small size can be crucial. 
The fundamental principle of the method is that the true underlying physical process $\mathbf{y}$ can be modeled by a model $\mathbf{f}$, tuned on a set of points $\mathbf{X}$ using a multivariate Gaussian distribution. This is the prior distribution, and without loss of generality, we can assume it is centered around zero. This assumption can be made more robust by standardizing the data so that their targets also fall within a zero mean Gaussian distribution. The covariance matrix can be defined by the use of a kernel function. A kernel is a distance-like user-chosen function that encodes the correlation of two outputs given the distance of their respective inputs. The prior would then take the mathematical form shown in
\begin{equation}\label{eq:prior}
  \mathbf{f}|\mathbf{X} \sim \mathcal{N}(0,\mathbf{K_{x,x}})
\end{equation}
where $\mathbf{K_{x,x}}$ is the kernel function applied pairwise to all input pairs to construct the covariance matrix.

The likelihood of the observations given the model can be represented as a noisy normal distribution around the model predictions:
\begin{equation}\label{eq:likelihood}
  \mathbf{y}|\mathbf{f} \sim \mathcal{N}(\mathbf{f},\mathbf{\sigma^2I})
\end{equation}
\Cref{eq:prior} and \Cref{eq:likelihood} can be combined through the Bayes formula (\Cref{eq:bayes}) to compute the posterior distribution which will serve as our prediction.
\begin{equation}\label{eq:bayes}
  P(\theta|\textbf{D}) =  \frac{P(\textbf{D} |\theta)P(\theta )}{P(\textbf{D})}.
\end{equation}
Solving the above process analytically parametrizes the posterior prediction and generates the prediction distribution.
\begin{subequations}
  \begin{align}
    \mathbf{\mu_{x^\star}}      & = \mathbf{K_{f^\star,f}[K_{f,f}+\sigma^2I]}^{-1}\mathbf{y} \label{eq:posterior_m}                                            \\
    \mathbf{\sigma_{x^\star}^2} & = \mathbf{K_{x^\star,x^\star}} - \mathbf{K_{x^\star,x}[K_{f,f}+\sigma^2I]}^{-1}\mathbf{K_{x,x^\star}} \label{eq:posterior_s}
  \end{align}
\end{subequations}
where the points for prediction and training are denoted with and without the superscript $\star$, respectively. 
Gaussian processes only require the definition of the prior mean and covariance matrix. The first can be routinely assumed to be zero, though other variations exist, such as a constant mean or a mean that linearly depends on the input values. The latter can be defined with a kernel function whose parameters should, in principle, be uncorrelated to the observations. However, it is common practice to optimize these parameters to maximize the marginal log likelihood to achieve better prediction results. 
Note that a complete mathematical analysis of Gaussian processes and their derivation is beyond the scope of this work. A comprehensive description can be found in Ref. \cite{williams2006}.

It must be noted that PYTHIA aims at providing predictions and optimization with a limited dataset. If the complexity of the system rises significantly, one must resort to the so-called batch GP \cite{batchgp}. These algorithms should be able to handle large datasets, though at the cost of somewhat decreased performance. However, with state of the art algorithms implemented in gpytorch \cite{gpfast1, gpfast2} Gps can easily handle datasets of a million samples with very limited approximations.

\subsection{Foundational Model}
Gaussian processes are a versatile family of algorithms. Nevertheless, they suffer significantly with feature spaces of large dimensionality. To this end, we have elected to process the structure using a Graph Neural Network (GNN). The complexity and sheer size of the chemical space needed to be investigated however makes the use of a simple GNN impractical. Therefore, we used an established and powerful foundational model, MACE \cite{MACE}. This is done without loss of generality and showcases the power of such architectures. Other architectures, such as NequIP \cite{NEQUIP} or potentially newer models can be seamlessly integrated.

\subsection{Translation Layer}

MACE is a powerful model, but its accuracy comes at a cost of high complexity and large parameter space. The typical size of datasets for adsorption application, however, does not suffice for the training of such a large model. Therefore, we are forced to utilize pre-trained models on adjacent properties and chemical species. Novel systems, however, differ significantly from the existing databases and models derived from these databases are not of-the-box applicable. For the model to be transferable, there is need for a translation layer that adapts the information extracted to the current case. 

The final output of the model however is too specific to use as all the general chemical information have been condensed to a target specific representation. To this end, we extract a deeper MACE layer, containing a rich molecular representation, to be used as input to the translation layer.  

The translation layer we have chosen is in the form of transformers. Transformers are a class of deep learning architectures designed to model complex dependencies in sequential or structured data using a mechanism known as self-attention. Introduced by Vaswani et al. (2017) \cite{vaswani2017attention}, the transformer architecture replaced traditional recurrent and convolutional models in many applications by enabling parallel processing of input data and capturing long-range relationships more effectively.

At the core of the transformer is the multi-head self-attention mechanism, which computes contextualized representations of each input token by attending to all other tokens in the sequence. Each attention head learns to focus on different types of relationships by projecting the input into query, key, and value vectors. The attention weights are calculated as scaled dot-products between queries and keys, and these weights are used to combine the value vectors, allowing the model to integrate information from relevant positions in the sequence.

Therefore, their ease of training on sparse data and their natural translation interpretation makes them suitable for transferring the knowledge of the foundational model on our very specific task and dataset. 

We do not need to retrain the MACE, a feat both computationally expensive and data inefficient. Instead we leverage its knowledge and only train the translation layer. The pipeline so far however lacks the element of uncertainty quantification. To add this, we use the translated vector as input to a Gaussian Processes model which provides the final prediction and the uncertainty quantification of this prediction. This allows us to leverage said predictions for Bayesian Optimization. 

\section{Results}
To evaluate PYTHIA we deployed it on two distinct cases, that of the Zn adsorption on the chini cluster family and the adsorption of formaldehyde (\ch{CH2O}) on the silicene. In both cases we optimize simultaneously for optimal adsorption position on adsorbent as well as for number and  type of defects in the chemical systems. We compare this approach with both the AUGUR framework as well as deploying the foundational model to be fully trained solely on the acquired dataset. 

\subsection{Zn adsorption on Chini Clusters}


Chini clusters are platinum (Pt) carbonyl clusters with a distinct chemical formula $[\mathrm{Pt}_3(\mathrm{CO})_6]_n^{2-}$, built from n stacked layers of triangular Pt units. These are widely used as precursors for synthesising atom-precise Pt nanoclusters (Pt-APNCs) which are obtained by encapsulating them inside a Zeolitic-imidazolate framework (ZIF-8) through a bottle-around ship approach \cite{Mechanistic_Poonam}.
ZIF-8 assembles from Zn$^{2+}$ nodes and 2-methylimidazolate linkers, so the chini cluster interacts with the Zn ions while the framework forms around it. We therefore used Zn adsorption energy as an optimization target, where more negative values denote stronger binding. The detailed methodology for computing energies is given in SI Section 2.1.

To generate APNCs with higher density of catalytically active sites, experimentalists have tried to substitute Pt with various metal atoms in the chini cluster geometry and have been successful \cite{PtNi_Chini_KathrinKratzl}.
Therefore, substitution of the metal atoms in the chini cluster architecture is a relevant design variable. We used Ni and Pd as substituents as both belong to group 10 and are isovalent with Pt. Substitution therefore preserves the valence electron count of the cluster and is expected to retain the chini framework, while changing the electronic structure that controls Zn binding. 
We considered clusters with $n = 1$ and $2$ and allowed any number of Pt atoms, up to all of them, to be substituted.
Each candidate structure was thus defined by its dopant configuration together with the Zn position, and PYTHIA optimises both jointly.

We assessed and compared three models for adsorption energy prediction: the AUGUR  baseline, MACE model trained fully with a GP regressor as a final layer, and PYTHIA, which combines a frozen pretrained MACE model with Transformers and Gaussian process. 

\begin{figure}[H]
    \centering
    \includegraphics[width=1\linewidth]{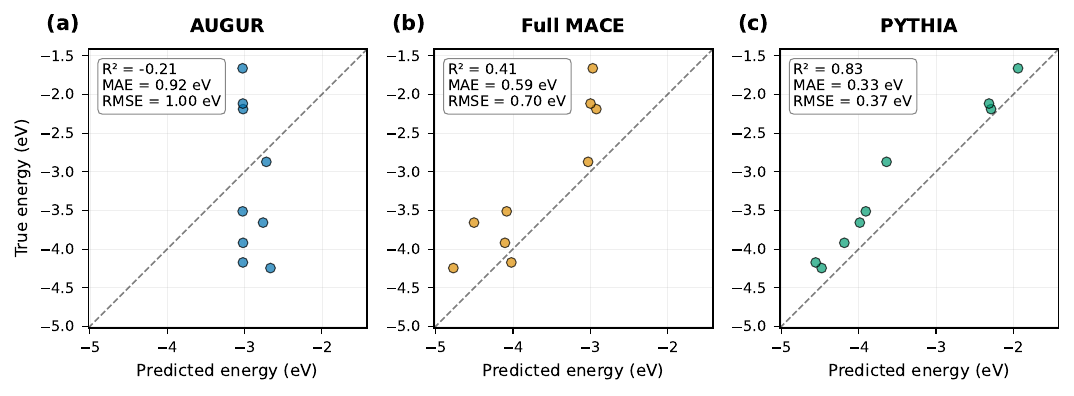}
    \caption{ Comparison of Zn$^{2+}$ adsorption energy predictions for doped Chini clusters a) a GNN with GP as the final layer, b) a MACE model trained fully with GP as a final layer c) PYTHIA. The points predicted correspond to systems with defect configuration not present in the training set.}
    \label{fig:performance_comparison}
\end{figure}

Figure~\ref{fig:performance_comparison} summarizes the relationship between predicted and reference adsorption energies. PYTHIA aligns most closely with the diagonal, returning an $R^2$ value of $0.83$, a mean absolute error (MAE) of $0.33$\,eV, and a root mean square error (RMSE) of $0.37$\,eV. This represents a significant improvement over the next best model, ranging between $80 \% - 100 \%$, depending on the chosen metric.

\begin{figure}[H]
    \centering
    \includegraphics[width=\linewidth]{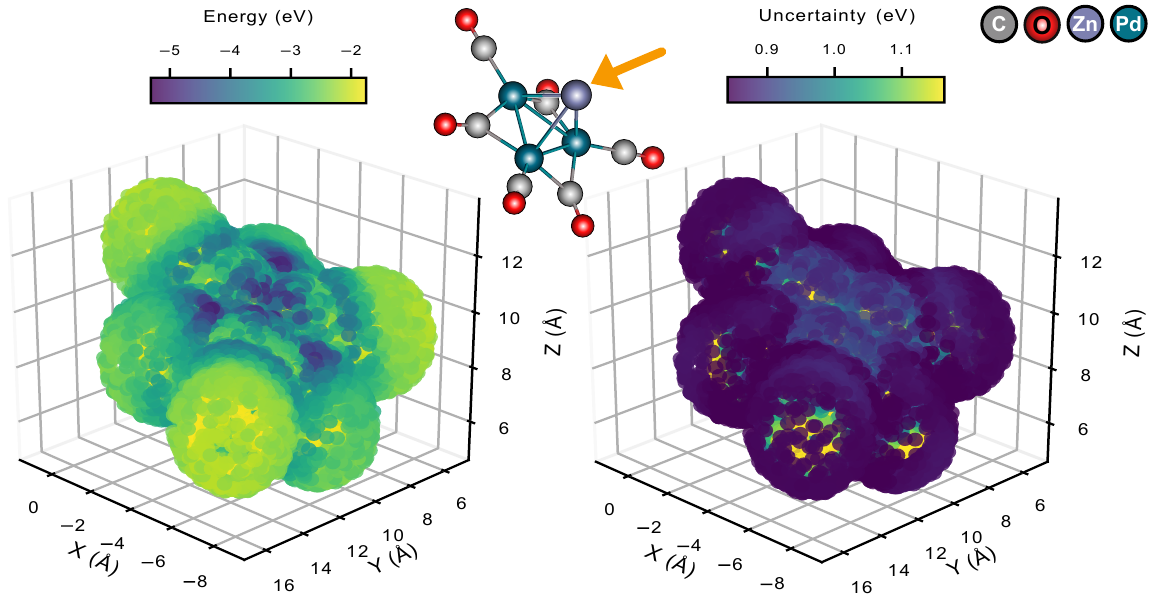}
    \caption{Energy landscape (left) and corresponding uncertainty (right) for the chini cluster (Pd$_3$, center) that exhibits the best Zn adsorption energy. As can be seen, the model uncertainty is uniformly low, which implies the model has acquired an understanding of the entire energy surface.}
    \label{fig:chini_energy}
\end{figure}

PYTHIA correctly captures the trends of the adsorption landscape of Zn adsorption on the chini clusters with respect to their defect engineering. Specifically, it identified highly promising dopants, and adsorption positions in the chini family. The predicted global minimum for this cluster lies at the  top or symmetric bottom sites with adsorption energy close to $-6.1$\,eV. It must be noted, that this optimal adsorption site was identified after just 1 optimization run, with the model being trained on a dataset that exhibited an optimum adsorption energy of $-4.74$ eV and did not contain Pd$_3$. PYTHIA provided an improvement of approximately $\mathbf{30\%}$ on the adsorption energy. Additionally, PYTHIA provides the full energy surface and an uncertainty quantification of any defective surface and can be seen in \cref{fig:chini_energy}. This allows for a deeper investigation of systems that range too far from the acquired model understanding and substitutes significant simulation times with minutes of predictions.

\subsection{Formaldehyde adsorption on Silicene}
The chini cluster provided a good baseline but the limited range of defects does not fully showcase the strength of PYTHIA. To this end, we apply our pipeline to a second system, that of formaldehyde adsorbed by defective silicene. This system allows for more diverse and higher number of defects on the cluster, with an added layer of complexity due to its periodic nature.. Additionally, the adsorbent has multiple atoms and needs to be optimized rotationally as well as positionally. 

Specifically, silicene was selected as the model substrate because its low-buckled structure and mixed sp$^2$/sp$^3$ bonding make it more chemically reactive than graphene \cite{massonEpitaxialGrowthStructural2023}.
This renders its adsorption properties highly sensitive to local structural and electronic modifications, driving considerable interest in silicene for chemical sensing and surface functionalization \cite{molleSiliceneSiliceneDerivatives2018}.
Formaldehyde (\ch{CH2O}) was chosen as the target adsorbate due to its importance as a hazardous volatile organic compound, for which sensitive room-temperature detection remains an active area of research \cite{joExclusiveUltrasensitiveDetection2021}.
To generate a chemically diverse optimization problem, we considered substitutional dopants spanning multiple classes with distinct electronic effects: isovalent Group IV elements (C, Ge), donor-type Group V elements (N, P, As), acceptor-type Group III elements (B, Al, Ga, In), and transition metals with partially filled d states.
Main-group dopants have been extensively investigated as a means of tailoring silicene's electronic structure, band gap, charge distribution, and adsorption behavior \cite{arumugamExploringElectrocatalyticPerformance2024,tiwariComparativeAnalysisBoron2023,teshomeEffectDopingControlling2017}, while transition-metal dopants introduce localized d-derived states that can give rise to magnetism, topological phases, and enhanced adsorbate interactions \cite{zhangQuantumAnomalousHall2013,kalpanadeviImprovingElectronicOptical2025,abdullahSpinpolarisedDFTModeling2021}.
Together, these dopant classes span a broad range of chemically distinct electronic environments, providing a diverse optimization landscape for the active-learning framework.
Finally, multiple simultaneous substitutions were included to further increase the complexity of the search space, allowing the framework to explore cooperative effects between different dopant species rather than being restricted to isolated substitutional defects.\\

\begin{figure}[H]
    \centering
    \includegraphics[width=1\linewidth]{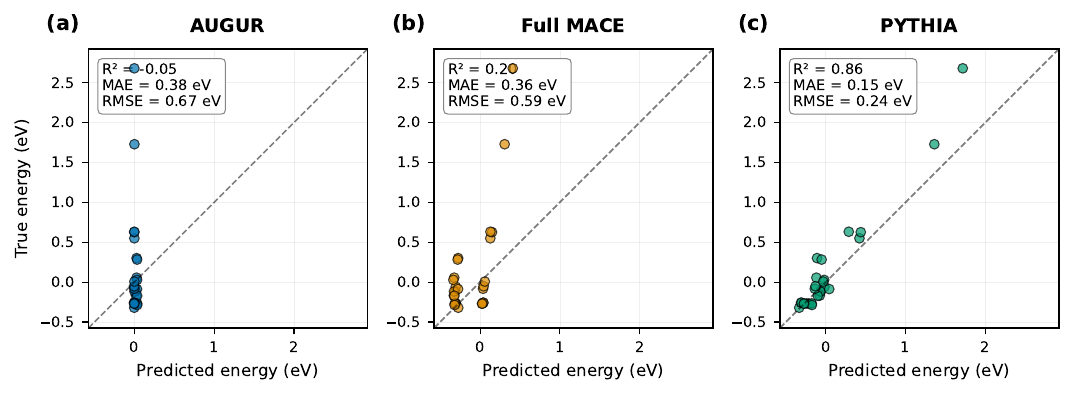}
    \caption{Comparison of adsorption energy prediction accuracy  for a defective Silicene-\ch{CH2O} system  for a) a GNN with GP as the final layer, b) a MACE model trained fully with GP as a final layer c) PYTHIA}
    \label{fig:si_comparison}
\end{figure}

The comparison between the baseline AUGUR model, fully trained MACE and PYTHIA
can be seen in \cref{fig:si_comparison}. Unlike the chini case, silicene has a much richer defect space and the adsorbate CH$_2$O more complex placement options.
It also follows that efficiency of PYTHIA will be much more prominently present. AUGUR fails entirely as the high complexity of the space cannot be easily captured with a simple graph neural network. In addition, the limited dataset does not allow for a full MACE training. Again, PYTHIA outperforms both approaches by a significant margin. 


We use the PYTHIA model to optimize both defects of adsorbent and position of adsorbate. For reasons of numerical stability
we restricted our defect optimization to maximum 3 defects in one adsorbent. This is not a limitation of PYTHIA and can be easily adjusted to fit a new case with more defects. 

\begin{figure}[H]
    \centering
    \includegraphics[width=\linewidth]{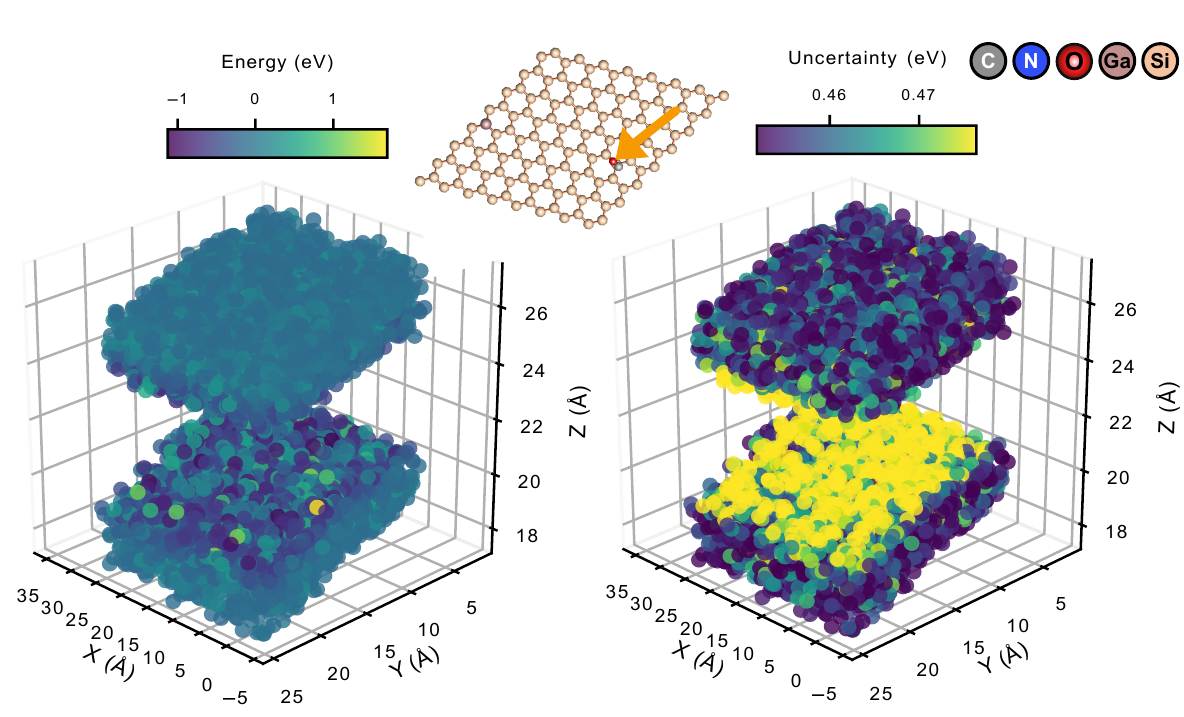}
    \caption{Energy landscape (left) and corresponding uncertainty (right) for the silicene that exhibits the best adsorption properties (Ga defect). The position of each energy point is that of the first Carbon of formaldehyde. Silicene is periodic so the energy surface is non closed. Further, formaldehyde can freely rotate and therefore each position can have multiple energies associated with it. As can be seen, the model uncertainty is uniformly low, which implies the model has acquired an understanding of the entire energy surface.}
    \label{fig:si_energy}
\end{figure}

The full energy surface of the best adsorbent can be found in \cref{fig:si_energy}. The reason for the relative discontinuity of the energy landscape is the fact that the adsorbate can be rotated, leading to many overlapping and wildly differing energies in comparatively close proximity. This, however, is only for visualization purposes and does not affect the optimization process. 
It must be noted that the position and defects determined have an energy of -1.12 eV, with only 2 iterations. The best interaction energy of the original dataset was -0.332 eV, with PYTHIA providing and improvement of more than $\mathbf{230\%}$. Further, the sheer effort of generating enough DFT calculations to fully explore the energy space now readily available would have taken many years of intense computational effort. PYTHIA however, provides this in minutes, allowing for a deeper understanding of the structure, far exceeding the mere optimum adsorption point determination. 

\section{Discussion}

In this work, we introduced PYTHIA (Probabilistic Yet Transferable Hierarchical Interaction Architecture), a data-efficient transferable machine learning framework for identifying optimal molecular interaction sites and defect choices for a variety of systems.  By combining a pre-trained, frozen foundational model (MACE) with a lightweight transformer-based translation layer and a Gaussian Process head, PYTHIA is able to achieve high accuracy with limited data, while simultaneously providing a robust uncertainty estimate. This makes PYTHIA suitable for a Bayesian Optimization framework, where efficient information acquisition is equally important to final accuracy. 
We have tested our pipeline in two highly relevant cases, namely Zn adsorption on the doped chini cluster family and CH$_2$O adsorption on doped silicene. Across both cases, PYTHIA outperformed all benchmarks leading to high improvement in the defect-site dual optimization problem, as well as resulting in much more accurate final models. It must be noted that the prediction accuracy of PYTHIA was high, even in defects not present in the dataset, thus paving the way to a full chemical space investigation, a luxury rarely afforded with conventional approaches.

This modularity, combined with the demonstrated transferability to systems and defect configurations not explicitly represented in the training set, suggests that PYTHIA is broadly applicable to other classes of interaction-site optimization problems in computational materials discovery — including catalyst design, gas sensing, and surface functionalization, where DFT data is scarce but adjacent, richly-labeled datasets exist to transfer from.
Lastly, PYTHIA is not limited by the case study or optimization objective. It can just as seamlessly be applied to determining ideal solvents as well as defects and adsorption sites. Additionally, as state of the art advances new models can seamlessly substitute MACE thus keeping this framework consistently relevant. 
The full pipeline will provide even more value in highly novel and complex materials, where large datasets do not exist and data accuracy is as vital as good optimization results. 

\section*{Data Availability}
Data will be made available via github upon acceptance for publication.
\section*{Code Availability}
Code will be made available via github upon acceptance for publication.

\section*{ACKNOWLEDGEMENTS}
 
  A.G. acknowledges financial support from TUM Innovation Network for Artificial Intelligence powered Multifunctional Material Design (ARTEMIS) and funding in the framework of Deutsche Forschungsgemeinschaft (DFG, German Research Foundation) under Germany's Excellence Strategy – EXC 2089/1 – 390776260 (e-conversion).
  Lastly, we wish to express our gratitude to Dr. Inigo Iribarren for creating the flowchart for this work.

   The Authors declare no Competing Financial or Non-Financial Interests.
\printbibliography

\end{document}